\documentclass[10pt,twocolumn,letterpaper]{article}

\usepackage{cvpr}
\usepackage{eso-pic}
\usepackage{times}
\usepackage{epsfig}
\usepackage{graphicx}
\usepackage{amsmath}
\usepackage{amssymb}
\usepackage{subcaption}
\usepackage[dvipsnames]{xcolor}
\usepackage{lastpage} 
\usepackage{booktabs}
\usepackage{authblk}

\definecolor{myblue}{RGB}{12,125,190}
\definecolor{myred}{RGB}{204,51,17}
\definecolor{mylightblue}{RGB}{51,187,238}
\definecolor{mypurple}{RGB}{240,103,153}
\definecolor{mygrey}{RGB}{187,187,187}
\definecolor{myorange}{RGB}{255,112,67}
\definecolor{mygreen}{RGB}{0,153,136}

\usepackage[pagebackref=true,breaklinks=true,letterpaper=true,colorlinks,bookmarks=false]{hyperref}

\cvprfinalcopy 

\begin{document}

\title{Enhance Multimodal Model Performance with Data Augmentation:\\ Facebook Hateful Meme Challenge Solution}

\author{
Yang Li
}
\author{
Zinc Zhang
}
\author{
Hutchin Huang
}
\affil{\textbf{Georgia Institute of Technology}}
\affil{{yli3297, zzhang889, hhuang400}@gatech.edu}

\maketitle

\begin{abstract}

Hateful content detection is one of the areas where deep learning can and should make a significant difference. The Hateful Memes Challenge from Facebook helps fulfill such potential by challenging the contestants to detect hateful speech in multi-modal memes using deep learning algorithms. In this paper, we utilize multi-modal, pre-trained models VilBERT and Visual BERT.  We improved models' performance by adding training datasets generated from data augmentation. Enlarging the training data set helped us get a more than 2\% boost in terms of AUROC with the Visual BERT model. Our approach achieved 0.7439 AUROC along with an accuracy of 0.7037 on the challenge's test set, which revealed remarkable progress. Our code is available at: \color{magenta}{\url{https://github.com/yangland/hatefulchallenge}}
   
\end{abstract}

\section{Introduction}

\begin{figure}[h]
    \centering
    \includegraphics[width=0.9\linewidth]{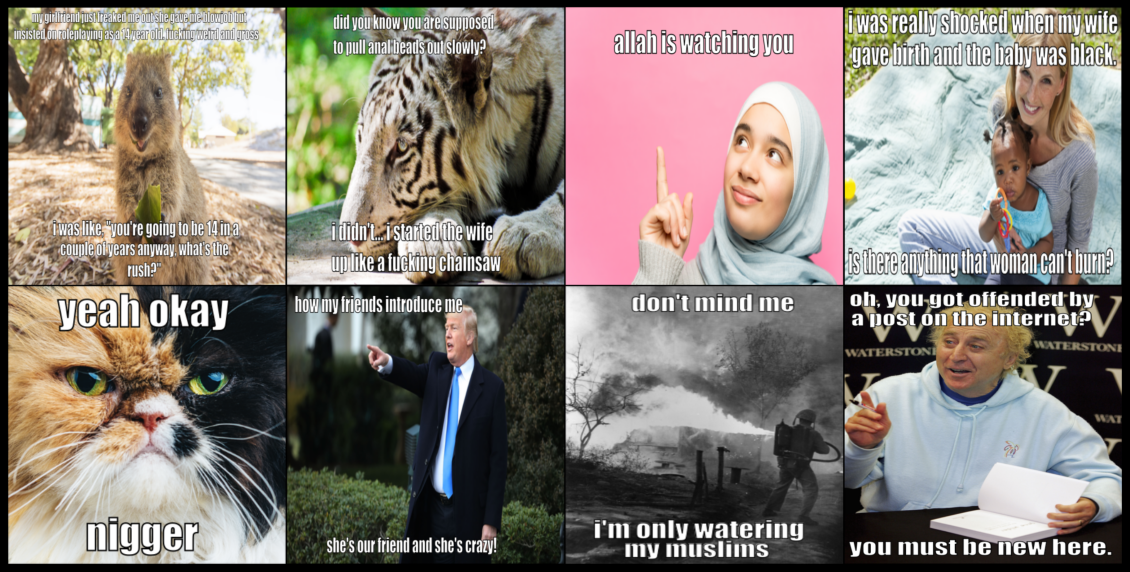}
    \caption{Hateful meme examples}
    \label{fig:example}
    \end{figure}

The Hateful Memes Challenges\cite{2005.04790} from Facebook introduces a data set where messages are made from both text and image. The traditional NLP and image processing methods have a hard time analyzing this type of memes due to their uni-modal nature. In this work, we investigate solutions for this challenge under Facebook's MMF framework, evaluate the performances of the multi-modal models and propose modal turning options. Example memes can be found in Figure \ref{fig:example}.

Despite the recent development for multi-modal learning from models such as ViLBERT\cite{1908.02265} and VisualBERT\cite{1908.03557}, the best model's performance is still far behind from average human performance\cite{2012.12871}. The difficulty of the hateful meme challenge is the knowledge for correct classification needs to be learned from another dataset. The hateful meme dataset is very small, with 10,000 meme records. It can only be used as examples and for validations. We can not use the hateful meme dataset as the only source for training. The commonsense and reasoning would require training from another more comprehensive dataset, which needs to have a broad coverage over characteristics that commonly been targeted in hateful posts such as sex, race, and religion. The multimodal learning model also needs to connect both images, natural language, and the attentions in-between to provide the support for classification to be successful. Since the representation model is task-agnostic, the model also requires further turning for the specific task in identifying Hateful Memes.

Visual Question Answering (VQA) and Visual Commonsense Reasoning (VCR) are tasks tested in the VisualBERT Research\cite{1908.03557}. They are related to the multimodal classification task for hateful meme challenges. These two tasks require cross-attention between text and image, and a background knowledge base that does not exist in the information from the current example.


Directly, our research could help social media companies such as Facebook and Twitter. A better automatic process could save the high cost of hiring human reviewers to identify inappropriate posts. On the other hand, a successful model could also guide publishers, media, and individual users, help them identifying potential inappropriate content before publishing. Broadly speaking, a better multi-modal model could sense more downstream task-related information, support better reasoning and apply better commonsense to visual-text contents.


We use the Hateful Memes dataset comes from the official Facebook Challenge website \href{https://www.drivendata.org/competitions/64/hateful-memes/page/206/}{here}. It contains in total 5 separate sets. One training set (train), two development sets (dev\_seen and dev\_unseen) and two test sets(test\_seen and test\_unseen). The training set has 8500 labeled samples. Amongst two development sets, dev\_seen has 500 labeled samples and dev\_unseen has 540 labeled samples. As for test sets, test\_seen and test\_unseen have 1000 and 2000 unlabeled samples respectively.

The data sets that we use as per-trained models are from the Facebook's MMF framework, including:
\begin{itemize}
  \item COCO\cite{1504.00325}
  \item COCO17
  \item Conceptual Captions\cite{sharma-etal-2018-conceptual}
  \item VQA2
\end{itemize}

\section{Method}


\begin{figure}[hbt!]
\centering
\includegraphics[width=1\linewidth]{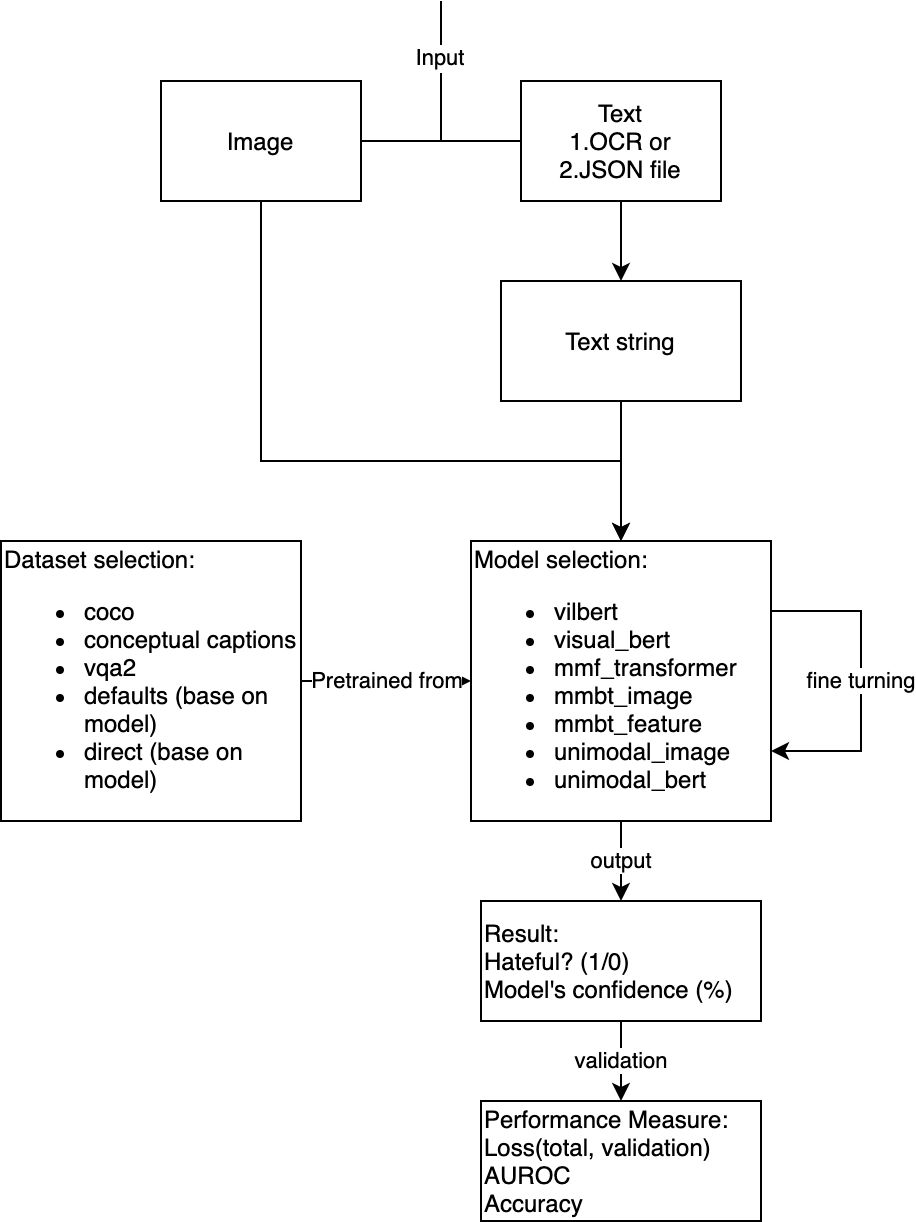}
\caption{Model screening workflow}
\label{fig:workflow}
\end{figure}
    
Our approach can be divided into two parts: model screening and fine tuning.

In part 1 of our approach, we implemented a quick survey over the seven models applicable to the Hateful Memes problem provided by the MMF framework’s “model zoo” (\url{https://mmf.sh/docs/notes/'model\_zoo'}), namely MMBT (“MMBT”), VilBERT (“ViLBERT”), Visual BERT(“Visual\_BERT”), MMF Transformer(“mmf\_transformer”), FUSIONS (“Concat\_BERT”, “Late\_Fusion”), and MMF\_BERT. To be specific, we decided to utilize these pre-trained models by conducting the same settings of training routine with a fixed epoch of 3,000 on the Hateful Memes dataset, and then compare these models with one another, qualitatively and quantitatively analyze their performance using both accuracy and AUROC. The workflow of model screening shows in Figure \ref{fig:workflow}.


Part 2 of our approach is to conduct a series of experiments to fine-tune the selected pre-trained models from the previous part. We would like to fine-tune the models from the following perspectives.

First, since the hateful meme itself is relatively small in observation numbers, we want to prevent the models from overfitting this data set. Hence, we decided to exploit the “early stop” option provided by the MMF framework. We plan to stop training immediately when the accuracy rate gets lower on the validation set.

Second, we want to try different loss functions other than Cross Entropy loss, such as focal loss \cite{focal_loss}, to reweight the observations in the dataset that our models have difficulty in classifying correctly. We learned from Goswami et al \cite{2005.04790} that “difficult examples (‘benign confounders’) are added to the dataset to make it hard to rely on unimodal signals”. We believe changing cross entropy loss to focal loss might help the model handle such “difficult examples”.




Even the MMF models have been per-trained, the process of loading and running the train and validation on the hateful meme data set still takes a significant amount of GPU and RAM resource. To overcome the physical challenge, our group uses a workstation equipped with a NVIDIA GeForce RTX 3090 GPU. Each of the experiments takes about 50 - 150 mins runtime under the setup of early stop. To save time, we freeze several front-end parts of the models. 
We consulted the MMF starting code from facebook's repository: \url{https://colab.research.google.com/github/facebookresearch/mmf/blob/notebooks/notebooks/mmf_hm_example.ipynb}


\section{Experiments and Results}


As we cannot access the labeled test dataset, and the competition has a daily submission limit, all measurements are done locally and only done on the development set. We chose dev\_unseen as the validation set. The model that achieves a higher ``score" on the validation set is considered better than other models. The ``score", under our configuration, can be validation accuracy or validation AUROC.

We select AUROC, the area under the receiver operating characteristic curve, as the measure of binary classification task's performance. Its formula is:
\begin{equation}
\mathrm{AUROC}=\int_{x=0}^{1} \mathrm{TPR}\left(\mathrm{FPR}^{-1}(x)\right) \mathrm{d}x
\end{equation}
\emph{TPR: true positive rate, }
\emph{FPR: false positive rate}

\subsection{Model Screening}
\label{model_screen}
\begin{figure*}[hbt!]

\begin{subfigure}{0.35\textwidth}
\includegraphics[width=0.9\linewidth]{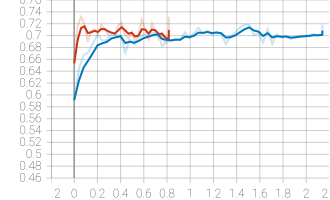} 
\caption{accuracy\\(\textcolor{myblue}{ViLBERT} vs \textcolor{myred}{Visual\_BERT})}
\label{fig_acc1}
\end{subfigure}
\begin{subfigure}{0.35\textwidth}
\includegraphics[width=0.9\linewidth]{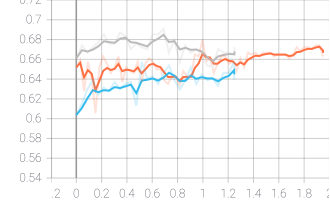}
\caption{accuracy\\(\textcolor{mylightblue}{MMF transformer} vs MMBT \textcolor{mygrey}{image} \textcolor{myorange}{features})}
\label{fig_acc2}
\end{subfigure}
\begin{subfigure}{0.35\textwidth}
\includegraphics[width=0.9\linewidth]{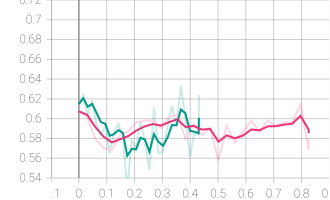}
\caption{accuracy\\(unimodal \textcolor{mygreen}{bert} vs \textcolor{mypurple}{image})}
\label{fig_acc3}
\end{subfigure}

\begin{subfigure}{0.35\textwidth}
\includegraphics[width=0.9\linewidth]{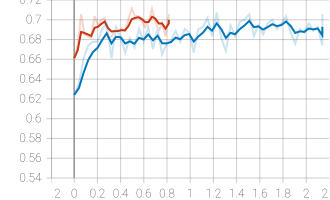} 
\caption{AUROC\\(\textcolor{myblue}{ViLBERT} vs \textcolor{myred}{Visual\_BERT})}
\label{fig_AUROC1}
\end{subfigure}
\begin{subfigure}{0.35\textwidth}
\includegraphics[width=0.9\linewidth]{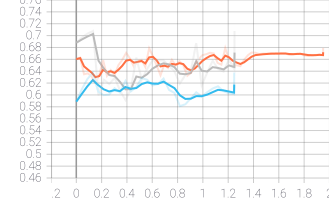}
\caption{AUROC\\(\textcolor{mylightblue}{MMF transformer} vs MMBT \textcolor{mygrey}{image} \textcolor{myorange}{features})}
\label{fig_AUROC2}
\end{subfigure}
\begin{subfigure}{0.35\textwidth}
\includegraphics[width=0.9\linewidth]{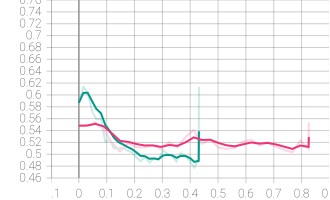}
\caption{AUROC\\(unimodal \textcolor{mygreen}{bert} vs \textcolor{mypurple}{image})}
\label{fig_AUROC3}

\end{subfigure}

\caption{Accuracy and AUROC for 7 pre-trained models}
\label{fig:7 pre-trained}
\end{figure*}

In the first round of screening, we tested seven pre-trained models, and their learning curves are present as three groups in Figure \ref{fig:7 pre-trained} for a clear drawing. We implemented an early stop function so the steps for experiments are different, but they all converged. The scores, including validation accuracy and validation AUROC, are listed in Table \ref{tab_model_selection}. We can see ViLBERT and Visual\_BERT performed best(\(\sim\) 0.72 AUROC), MMBT family the second (\(\sim\) 0.7 AUROC), and the unimodal family worst (\(\sim\)  0.6 AUROC).

We can see that the unimodal family did not perform as well as the rest of the models. Unimodal-image is an image-only modal based on ResNet152, and unimodal-bert is a text-only modal based on text transformer. The meme is made from both text and image, it is easy to understand why the unimodal models which only focusing on either text or image couldn't achieve better results. Observing the samples from the Hateful Memes dataset, we can find that many ``hateful" meanings are expressed through antonyms, satire, or puns between the meaning of the text and the semantics of the image. Unluckily, the unimodal models are lacking the ability to incorporate the information in the other modal and create a cross-attention relationship between them.

MMBT(Multimodal Bitransformers)\cite{1909.02950} made a significant improvement over the performance compare with the unimodal thanks to its BERT-like multimodal architectures. They leveled up AUROC by about 0.1, and accuracy by about 0.05. The advantage of these structures is to employ self-attention over both text and image modalities simultaneously. The result is an easier and more fine-grained multi-modal fusion. MMBT's architecture components are pre-trained individually as a unimodal task, while its internal structure allows information from different modalities to interact at multiple different levels via self-attention instead of just the final layer. Compare with ViLBERT, MMBT is simpler, but the ViLBERT's BERT architecture is trained on the multi-model dataset. This could explain why in the final result ViLBERT is still slightly better than MMBT.

Our best candidates are ViLBERT and Visual\_BERT, who achieved AUROC of 0.7187 and 0.7319 respectively. ViLBERT is a BERT architecture that extends to a multi-modal two-steam model. ViLBERT's structure is designed for image content and natural language. Per-trained with a large dataset like Conceptual Captions, ViLBERT can perform well on vision-and-language tasks including Visual Question Answering (VQA) and Visual Commonsense Reasoning (VCR). The knowledge and commonsense were learned as a semantically meaningful alignment between vision and language during pretraining. Another approach, VisualBERT incorporates BERT with pre-trained object proposals systems such as Faster-RCNN\cite{1908.03557}. Similar to ViLBERT, VisualBERT also has multiple Transformer layers that jointly process the text and image inputs. Different from ViLBERT, VisualBERT uses sets of visual embeddings to pass object information from image input to multi-layer Transformer along with the original set of text embeddings. Pre-trained with large data set such as COCO, VisualBERT performance well on VQA, VCR, NLVR\_2, and Flicker30K tasks.

The reasons that ViLBERT and Visual\_BERT performed better in the Hateful Memes' classification task could be 1) Their multimodal per-training. 2) Their ability to use multiple transformers layers to closely connect image and text inputs. For both ViLBERT and VisualBERT, increasing the size of the pretraining data set or incorporate more data sets that are related to the downstream application could potentially enhance the performance.


\begin{table}[h]
\centering
\caption{Scores in Model Selection Phase}
\label{tab_model_selection}
\begin{tabular}{ccc}
\toprule
& Accuracy & AUROC \\
\midrule
MMF Transformer & 0.6389 & 0.6377  \\
MMBT Image & 0.6722 & 0.7078 \\
MMBT Features & 0.6630 & 0.6791 \\
Unimodal Image & 0.5796 & 0.5539 \\
Unimodal Bert  & 0.6241 & 0.6135 \\
ViLBERT  & 0.7056 & 0.7187 \\
Visual BERT  & 0.7056 & 0.7319 \\
\bottomrule
\end{tabular}
\end{table}

\subsection{Fine Tuning}
\label{fine_tuning}

Fine tuning phase was conducted on the models that were passed in model screen phase, which were ViLBERT and Visual\_BERT as mentioned in \ref{model_screen}. We implemented two ways to tune each of two models: 1) Use Focal loss rather than naive binary cross entropy loss; 2) Enlarge the train set by adding new labeled samples. These two ways are conducted separately, but both of them are used along with early stop enabled.

Focal loss need two predefined hyper parameters: $\alpha$ and $\gamma$. $\alpha$ controls the weight that balances the ratio between positive and negative samples. When the amount of negative samples is dominant, a smaller $\alpha$ is preferred to lower the impact by negative samples. $\gamma$ is the focusing parameter, a weight to adjust the training strength on hard-to-classify samples. In our configuration, $\gamma$ is set to 2.0 according to \cite{focal_loss}. As for $\alpha$, we found that the positive samples account for about 30\% of the dataset, which should not be considered an imbalance of samples, therefore we simply set $\alpha$ to 1.0 empirically.

One thing worthing to note is that in Figure \ref{fig:7 pre-trained}, all seven models reached their performance peak within just one or two thousand iterations, and then gradually corrupted afterward. It suggests that the training set is somehow small. Regular fine-tuning on them will fast saturate. Without merged with the data sets that the models pre-trained on, they cannot contain the knowledge they already have for a long time and soon lose this knowledge. To relieve this problem, using an augmented training set is a practical way in fine-tuning phase.

However, enlarging the training set is simple but a little tricky. The safest way to obtain new data is to collect a large amount of memes with or without text from the Internet, and then manually label them one by one. This method should be the most reliable, but it will take a long time. Therefore, we adopted an alternative scheme to increase the data set. First, since we only do validation on dev\_unseen, and dev\_seen contains additional annotation data, we merge the data in dev\_seen that does not appear in dev\_unseen into the training set, a total of 100 samples. Second, we used keywords such as ``sarcasm memes", ``irony memes", and ``scorn memes" to collect 1500 additional pictures with the text through Google Image and Pinterest, and read the text on memes through Huawei Cloud OCR. It makes additional data consistent with the format of the hateful meme. To let the new data be compatible with our models, we discarded the images with any non-alphanumeric characters. After this, 1147 images were kept. Although without thorough manual cleaning, we believe that among the memes obtained through the keywords mentioned above, those that were detected with texts by OCR can be marked as ``hateful", and those that were detected without any text are all labeled as ``not hateful". In the end, we got an expanded training data set with 8500+100+1147=9747 labeled samples.

\begin{table}[ht]
\centering
\caption{Scores in Fine Tuning Phase}
\label{tab_fine_tune}
\begin{tabular}{ccc}
\toprule
& Accuracy & AUROC \\
\midrule
ViLBERT  & 0.7056 & 0.7187 \\
ViLBERT w/ Focal Loss & 0.6852 & 0.7188 \\
ViLBERT w/ more data & 0.7019 & 0.7227 \\
\midrule
Visual BERT  & 0.7056 & 0.7319 \\
Visual BERT w/ Focal Loss & 0.7056 & 0.7070 \\
Visual BERT w/ more data & 0.7037 & 0.7439 \\
\bottomrule
\end{tabular}
\end{table}

\begin{figure*}[h]

\begin{subfigure}{0.5\textwidth}
\includegraphics[width=1\linewidth]{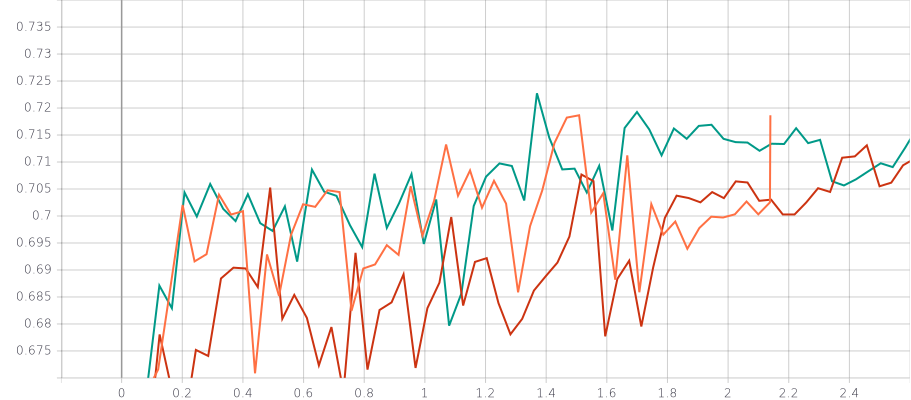} 
\caption{Vilbert \textcolor{mygreen}{extral}, \textcolor{myred}{focalloss}, \textcolor{myorange}{original}}
\label{fig_ft_acc1}
\end{subfigure}
\begin{subfigure}{0.5\textwidth}
\includegraphics[width=1\linewidth]{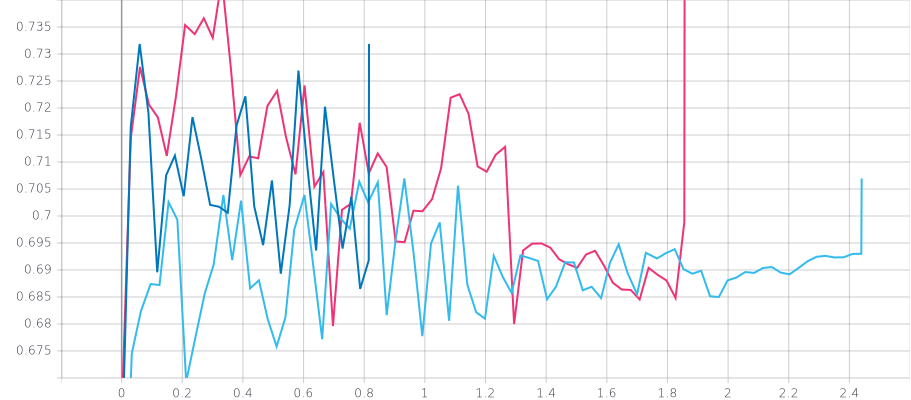}
\caption{Visual\_BERT \textcolor{mypurple}{extral}, \textcolor{mylightblue}{focalloss}, \textcolor{myblue}{original}}
\label{fig_ft_acc2}
\end{subfigure}

\caption{AUROC for Vilbert(left) and Visual\_BERT(right)}
\label{fig:Vilbert vs. Visual_BERT}
\end{figure*}

The results of fine tuning phase are listed in Table \ref{tab_fine_tune}. Their learning curves are shown in Figure \ref{fig:Vilbert vs. Visual_BERT}

Using Focal Loss, unfortunately, failed to improve VilBERT and Visual BERT. Although ViLBERT achieved a 0.0001 improvement in AUROC with the application of Focal Loss, this degree of improvement can be considered as random error. This shows that the application of Focal Loss may not be suitable for the Hateful Memes data set. Focal Loss was originally used for object detection tasks. The characteristic of object detection is that the number of negative samples is much greater than the number of positive samples, so it should be effective to reweight them for difficult samples. In the Hateful Memes data set, as mentioned above, the number of positive samples is more than 30\%. In this case, Focal Loss makes the model get an inappropriate reweighted gradient, making its performance on the validation set worse.

The use of the expanded data sets improves the performance of both ViLBERT and Visual BERT. Visual BERT's AUROC has increased by more than two percentage points. ViLBERT's AUROC has increased by nearly half a percentage point. However, the accuracy of both has a slight decrease. Overall, the fine tuning strategy using augmented data sets is effective. This also confirms our previous analysis on the learning curve. Using a larger data set can make the fine tune model more robust. Though from Figure \ref{fig:Vilbert vs. Visual_BERT}, we can see that overfitting still exists, however is slightly delayed compared to those in Figure \ref{fig:7 pre-trained}

\begin{figure}[hbt!]
\centering
\includegraphics[width=0.4\linewidth]{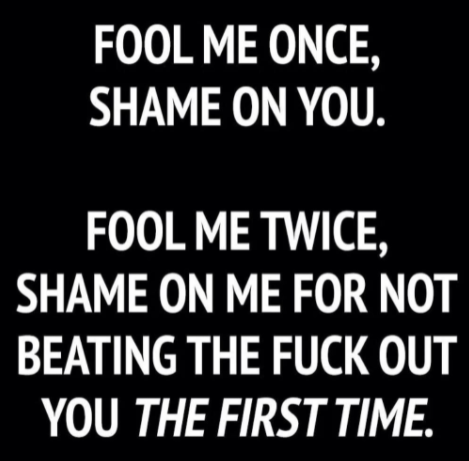}
\caption{A new-added sample with pure black background}
\label{fig:extra_example}
\end{figure}


As for why the accuracy of both has decreased, we think that there are some of the new-added samples, as shown in Figure \ref{fig:extra_example}, are pure black backgrounds with some texts. Compared with the original training set and development set, this sample is very much different. Such samples may harm the performance of the model.






\section{Conclusion and Future Work}


We proposed an approach detecting Hateful Memes in a small multimodal dataset from Facebook. We set up a common workflow to exam and measure the performance of different models. We conclude that Visual\_BERT and ViLBERT are better models for this challenge. We adjusted our fine-turn strategy and proved data augmentation is an effective solution for both Visual\_BERT and ViLBERT. Our approach achieves 0.7439 AUROC with an accuracy of 0.7037 on the challenge test set. We hope that these observations can further inform future research in the field.

The combination of text and images is a huge sample space, so a large number of samples are essential. ImageNet\cite{image_net} is a well-known image classification pre-training data set with over one million samples. The corpus used by BERT has reached the magnitude of ten million. In contrast, the amount of data contained in Hateful Memes is too small. Future work can start from the amount of data, take into account samples with multiple characteristics, and use automatic or semi-automatic methods to increase the number of data sets at least several times the original.

Another direction is data augmentation on the hateful-meme dataset. Some color-related operations, such as color jitter, inversion, channel swap, or deblurring, might help the model get out from being trapped object-unrelated details in the images. Shape-related techniques like rotation and perspective transforming might get rid of some biased information from the images. Also, recent GAN-based techniques, like SRNet\cite{srnet}, can directly replace the text on an image with a new one while keeping its original style. This technique could augment the data by giving us a lot of edited memes with any labels we want. Through those augmentations can the data set for surely be enlarged for a considerable amount, and our model can finally benefit from that.

\newpage

{\small
\bibliographystyle{ieee_fullname}
\bibliography{main}

\begin{thebibliography}{10}\itemsep=-1pt

\bibitem{1504.00325}
Xinlei Chen, Hao Fang, Tsung-Yi Lin, Ramakrishna Vedantam, Saurabh Gupta, Piotr
  Dollar, and C.~Lawrence Zitnick.
\newblock Microsoft coco captions: Data collection and evaluation server, 2015.

\bibitem{1909.02950}
Douwe Kiela, Suvrat Bhooshan, Hamed Firooz, Ethan Perez, and Davide Testuggine.
\newblock Supervised multimodal bitransformers for classifying images and text,
  2019.

\bibitem{2005.04790}
Douwe Kiela, Hamed Firooz, Aravind Mohan, Vedanuj Goswami, Amanpreet Singh,
  Pratik Ringshia, and Davide Testuggine.
\newblock The hateful memes challenge: Detecting hate speech in multimodal
  memes, 2020.

\bibitem{1908.03557}
Liunian~Harold Li, Mark Yatskar, Da Yin, Cho-Jui Hsieh, and Kai-Wei Chang.
\newblock Visualbert: A simple and performant baseline for vision and language,
  2019.

\bibitem{focal_loss}
Tsung-Yi Lin, Priya Goyal, Kaiming~He Ross~Girshick, and Piotr Dollár.
\newblock Focal loss for dense object detection, 2017.

\bibitem{2012.12871}
Phillip Lippe, Nithin Holla, Shantanu Chandra, Santhosh Rajamanickam, Georgios
  Antoniou, Ekaterina Shutova, and Helen Yannakoudakis.
\newblock A multimodal framework for the detection of hateful memes, 2020.

\bibitem{1908.02265}
Jiasen Lu, Dhruv Batra, Devi Parikh, and Stefan Lee.
\newblock Vilbert: Pretraining task-agnostic visiolinguistic representations
  for vision-and-language tasks, 2019.

\bibitem{image_net}
Olga Russakovsky, Jia Deng, Hao Su, Jonathan Krause, Sanjeev Satheesh, Sean Ma,
  Zhiheng Huang, Andrej Karpathy, Aditya Khosla, Michael Bernstein,
  Alexander~C. Berg, and Li Fei-Fei.
\newblock Imagenet large-scale visual recognition challenge, 2014.

\bibitem{sharma-etal-2018-conceptual}
Piyush Sharma, Nan Ding, Sebastian Goodman, and Radu Soricut.
\newblock Conceptual captions: A cleaned, hypernymed, image alt-text dataset
  for automatic image captioning, July 2018.

\bibitem{srnet}
Liang Wu, Chengquan Zhang, Jiaming Liu, Junyu Han, Jingtuo Liu, Errui Ding, and
  Xiang Bai.
\newblock Editing text in the wild, 2019.

\end{thebibliography}
}

\end{document}